\documentclass{book}
\usepackage{chicagoc}
\usepackage{kcp1}

\newcommand{\wbox}{\mbox{$\sqcap$\llap{$\sqcup$}}}
\newcommand{\xam}{\begin{example}}
\newcommand{\exam}{\end{example}}
\newcommand{\dfn}{\begin{definition}}
\newcommand{\edfn}{\end{definition}}

\collection
\newcommand{\commentout}[1]{}
\begin{document}
\newcommand{\ML}{\mathit{ML}}
\newcommand{\FF}{F}
\newcommand{\FS}{\mathit{FS}}
\newcommand{\BH}{\mathit{BH}}
\newcommand{\nBS}{\mathit{BS}}
\newcommand{\BT}{\mathit{BT}}
\newcommand{\SH}{\mathit{SH}}
\newcommand{\ST}{\mathit{ST}}
\newcommand{\LT}{\mathit{LT}}
\newcommand{\RT}{\mathit{RT}}
\newcommand{\LB}{\mathit{LB}}
\newcommand{\RB}{\mathit{RB}}
\newcommand{\BS}{\mathit{BS}}
\newcommand{\MT}{\mathit{T}}
\newcommand{\F}{{\cal F}}
\renewcommand{\S}{{\cal S}}
\newcommand{\U}{{\cal U}}
\newcommand{\V}{{\cal V}}
\newcommand{\VS}{\mathit{VS}}
\newcommand{\R}{{\cal R}}
\newcommand{\union}{\cup}

\commentout{
}

\newcommand{\sat}{\models}

\setcounter{chapter}{22}
\paper[Actual Causation and the Art of Modeling]{Actual Causation and the
Art of Modeling}{Joseph Y. Halpern and Christopher Hitchcock}

\commentout{
\author{
Joseph Y. Halpern\thanks{Supported in part by NSF grants 
IIS-0534064 and IIS-0812045, and by AFOSR grants
FA9550-08-1-0438 and FA9550-05-1-0055.}
\\
Cornell University\\
halpern@cs.cornell.edu
\and
Christopher Hitchcock\\
California Institute of Technology\\
cricky@caltech.edu}
}


\section{Introduction}

In \emph{The Graduate}, Benjamin Braddock (Dustin Hoffman) is told that
the future can be summed up in one word:  ``Plastics''.  One of us
(Halpern) recalls that in roughly 1990, Judea Pearl told him that the
future was in causality.  Pearl's own research was largely focused on
causality in the years after that; his seminal contributions are widely
known. We were among the many influenced by his work.  We discuss one
aspect of it, \emph{actual causation}, in this article,
although a
number of our comments apply to causal modeling more generally.  

Pearl introduced a novel account of actual causation in Chapter 10 of
\emph{Causality}, which was later revised in collaboration with one of
us \cite{HP01b}.
\nocite{pearl:2k}
%
In some ways, Pearl's approach to actual causation can be seen as a
contribution to the philosophical project of trying to analyze actual
causation in terms of counterfactuals, a project associated most
strongly with David Lewis \citeyear{Lewis73a}. But Pearl's account was
novel in at least two important ways. 
The first was his use of \emph{structural equations} as a tool for
modeling causality.  
In the philosophical 
literature, causal structures were often represented using so-called
\emph{neuron diagrams},   
but these are not (and were never intended to be) all-purpose
representational tools. 
(See \cite{Hitchcock07b} for a detailed discussion of the
limitations of neuron diagrams.)
We believe that the lack of a more 
adequate representational tool had been a serious obstacle to
progress.     
Second, while the philosophical literature on causality has focused almost
exclusively on actual causality, for Pearl, actual causation was 
a rather specialized topic within the study of causation,
peripheral to many issues involving causal reasoning and inference.
Thus, Pearl's work placed the study of actual causation within a much
broader context. 

The use of structural equations as a model for causal relationships
was well known long before Pearl came on the scene; it seems to go back
to the work of Sewall Wright in the 1920s (see \cite{Goldberger72} for a
discussion).  However, the details of the framework that have proved so
influential are due to Pearl.  
Besides the Halpern-Pearl approach mentioned above, 
there have been a number of other closely-related approaches for using  
structural equations to model actual causation; see, for
example, \cite{GW07,Hall07,hitchcock:99,Hitchcock07,Woodward03}.
The goal of this paper is to look more carefully at the modeling of
causality using structural equations.
For definiteness,  we use the Halpern-Pearl (HP) version 
\cite{HP01b} here, but our comments apply equally well to the other 
variants.

It is clear that the structural equations can have a major impact on the
conclusions we draw about causality---it is the equations that allow us
to conclude that lower air pressure is the cause of the lower barometer
reading, and not the other way around; increasing the barometer reading
will not result in higher air 
pressure.  The structural equations express the 
effects 
of
\emph{interventions}: what happens to the bottle if it is
hit with a hammer; what happens to a patient if she is treated with a
high dose of the drug, and so on.  These effects are, in principle,
objective; the structural equations can be viewed as describing
objective features of the world.  However, as pointed out 
by Halpern and Pearl \citeyear{HP01b} and reiterated by others 
\cite{Hall07,hitchcock:99,Hitchcock07}, the choice of 
variables and their values can also have a significant impact on
causality.  Moreover, these choices are, to some extent, subjective.
This, in turn, means that judgments of actual causation are subjective.

Our view of actual causation being at least partly subjective stands in
contrast to the 
prevailing view in the philosophy literature, where 
the assumption is that the job of the philosopher is to analyze
the (objective) notion of causation, rather like that of a chemist
analyzing the structure of a molecule.    
This may stem, at least in part, from failing to appreciate one of
Pearl's lessons: 
actual causality is only part of the bigger picture of causality.  There
can be an element of subjectivity in ascriptions of actual causality
without causation itself being completely subjective.   In any case,
the experimental evidence
certainly suggests that people's views of causality are subjective, even
when there is no disagreement about the relevant structural
equations. For example, a number of experiments 
show that broadly normative considerations, including the subject's own
moral beliefs, affect causal judgment. 
(See, for example, \cite{Alicke92,Cushman09,CKS08,HK09,KF08}.)
Even in relatively non-controversial cases, people may want to focus on
different aspects of a problem, and thus give different answers to
questions about causality.  For example, suppose that we ask for the cause 
of a serious traffic accident.  A traffic engineer might say that the bad
road design was the cause; an educator might focus on poor driver's
education; a sociologist might point to the pub near the highway where
the driver got drunk; a psychologist might say that the cause is the
driver's recent breakup with his girlfriend.\footnote{This is a 
variant of an example originally due to 
Hanson \citeyear{Hanson58}.} 
Each of these answers is
reasonable.  By appropriately choosing the variables, the structural
equations framework can accommodate them all.

Note that we said above ``by appropriately choosing the variables''.  An
obvious question is ``What counts as an appropriate choice?''.  More
generally, what makes a model an appropriate model?
While we do want to allow for subjectivity, we need to be able
to justify the modeling choices made.  A lawyer in court trying to argue
that faulty brakes were the cause of the accident needs to be able to
justify his model; similarly, his opponent will need to understand what
counts as a legitimate attack on the model.  
In this paper
we discuss 
what we
believe are reasonable bases for such justifications.  Issues such as
model stability and interactions between the events corresponding to
variables turn out to be important.

Another focus of the paper is the use of defaults in causal reasoning.
As we hinted above, the basic structural equations model does not seem
to suffice to completely capture all aspects of causal reasoning.  To
explain why, we need to briefly outline how actual causality is defined
in the structural equations framework.
Like many other definitions of causality (see, for example,
\cite{Hume39,Lewis73}), the HP definition is based on counterfactual
dependence.  Roughly speaking, $A$ is a cause of $B$ if, had $A$ not
happened (this is the counterfactual condition, since $A$ did in fact
happen) then $B$ would not have happened.  As is well known, this naive
definition does not capture all the subtleties involved with causality.
Consider the following example (due to Hall \citeyear{Hall98}): 
Suzy and Billy both pick up rocks and throw them at  a bottle.
Suzy's rock gets there first, shattering the
bottle.  Since both throws are perfectly accurate, Billy's would have
shattered the bottle had Suzy not thrown.
Thus, according to the naive counterfactual 
definition, Suzy's throw is not a cause of the bottle shattering.
This certainly seems counterintuitive.

The HP definition deals with this problem by taking $A$ to be a cause of
$B$ if $B$ counterfactually 
depends on 
$A$
{\em under some contingency}.  For example, Suzy's throw
is the cause of the bottle 
shattering because the bottle shattering counterfactually depends on
Suzy's throw, under the contingency that Billy doesn't throw.
(As we will see below, there are further subtleties in the definition that guarantee that, if
things are modeled appropriately, Billy's throw is not also a cause.)

While the definition of 
actual causation in terms of structural equations has
been successful at dealing with many of the problems of causality, 
examples of Hall \citeyear{Hall07}, Hiddleston
\citeyear{Hiddleston05}, and Hitchcock \citeyear{Hitchcock07} show that it
gives  
inappropriate answers in cases that have structural equations isomorphic
to ones where it arguably gives the appropriate answer.
This means that, no matter how we define actual causality in the
structural-equations framework, the definition must involve more than 
just the structural equations.  Recently,
Hall \citeyear{Hall07}, Halpern \citeyear{Hal39}, and 
Hitchcock \citeyear{Hitchcock07}
have suggested
that using
defaults might be a way of dealing with the problem.  As the
psychologists Kahneman and
Miller \citeyear[p.~143]{KM86} observe,
``an event is more likely to be undone by altering exceptional than
routine aspects of the causal chain that led to it''.  
This intuition is also present in the legal literature.  Hart and Honor{\'e}
\citeyear{HH85} observe that the statement ``It was the presence of
oxygen that caused the fire'' makes sense only if there were reasons to
view the presence of oxygen as abnormal.  

As shown by Halpern \citeyear{Hal39}, we can model this intuition
formally by combining a well-known approach to modeling defaults and
normality, due to Kraus, Lehmann, and Magidor \citeyear{KLM} with the
structural-equation model.  Moreover, doing this leads to a
straightforward solution to the problem above.  
The idea is that, when showing that if $A$
hadn't happened then $B$ would not have happened, we consider only
contingencies that are 
at least as
normal 
as
the actual world.
For example, if someone typically leaves work at 5:30 PM and arrives home at
6, but, due to unusually bad traffic, arrives home at 6:10, the bad
traffic is typically viewed as the cause of his being late, not 
the fact that he left at 5:30 (rather than 5:20).  

But once we add defaults to the model, the problem of justifying the
model becomes even more acute.  We not only have to justify the structural
equations and the choice of variables, but also the default theory.  The
problem is exacerbated by the fact that default and ``normality'' have
a number of interpretations.  Among other things, they can represent
moral obligations, societal conventions, prototypicality information,
and statistical information.  All of these interpretations are relevant
to understanding causality; this makes justifying default choices
somewhat subtle.

The rest of this paper is organized as follows.
In Sections~\ref{sec:causalmodel} and~\ref{sec:actcaus}, we review the
notion of causal model and the HP definition of actual cause; most of
this material is taken from \cite{HP01b}.  In
Section~\ref{sec:variables}, we discuss some issues involved in the
choice of variables in a model.  In Section~\ref{sec:final}, we review
the approach of \cite{Hal39} for adding considerations of normality to
the HP framework, and discuss some modeling issues that arise when we do
so.  We conclude in Section~\ref{sec:conc}.


\section{Causal Models}\label{sec:causalmodel}

In this section, we briefly review the HP definition of causality.
The description of causal models given here is taken from
\cite{Hal39}, which in turn is based on that of \cite{HP01b}.

The HP approach assumes that the world is described in terms of random
variables and their values.  
For example, if we are trying to determine
whether a forest fire was caused by lightning or an arsonist, we 
can take the world to be described by three random variables:
\begin{itemize}
\item 
$\FF$ for forest fire, where $\FF=1$ if there is a forest fire and
$\FF=0$ otherwise; 
\item
$L$ for lightning, where $L=1$ if lightning occurred and $L=0$ otherwise;
\item
$\ML$ for match (dropped by arsonist), where $\ML=1$ if the arsonist
drops a lit match, and $\ML = 0$ otherwise.
\end{itemize}

Some random variables may have a causal influence on others. This
influence is modeled by a set of {\em structural equations}.
For example, 
to model the fact that
if either a match is lit or lightning strikes, then a fire starts, we could use
the random variables $\ML$, $\FF$, and $L$ as above, with the equation
$\FF = \max(L,\ML)$. (Alternately, if a fire requires both causes to be
present, the equation for $\FF$ becomes $\FF = \min(L,\ML)$.) 
The equality sign in this equation should be thought of more like an 
assignment statement in programming languages; once we set the values of
$\FF$ and $L$, then the value of $\FF$ is set to their maximum.  However,
despite the equality, if a forest fire starts some other way, that does not
force the value of either $\ML$ or $L$ to be 1.  


It is conceptually useful to split the random variables into two
sets: the {\em exogenous\/} variables, whose values are
determined by 
factors outside the model, and the
{\em endogenous\/} variables, whose values are ultimately determined by
the exogenous variables.  For example, in the forest-fire example, the
variables $\ML$, $L$, and $\FF$ are endogenous. 
However, we want to take as given that there is
enough oxygen for the fire and that the wood is sufficiently dry to
burn.  In addition, we do not want to concern ourselves with the factors
that make the arsonist drop the match or the factors that cause lightning.
These factors are all determined by the exogenous variables.  

Formally, a \emph{causal model} $M$
is a pair $(\S,\F)$, where $\S$ is a \emph{signature}, which explicitly
lists the endogenous and exogenous variables  and characterizes
their possible values, and $\F$ defines a set of \emph{modifiable
structural equations}, relating the values of the variables.  
A signature $\S$ is a tuple $(\U,\V,\R)$, where $\U$ is a set of
exogenous variables, $\V$ is a set 
of endogenous variables, and $\R$ associates with every variable $Y \in 
\U \union \V$ a nonempty set $\R(Y)$ of possible values for 
$Y$ (that is, the set of values over which $Y$ {\em ranges}).
$\F$ associates with each endogenous variable $X \in \V$ a
function denoted $F_X$ such that $F_X: (\times_{U \in \U} \R(U))
\times (\times_{Y \in \V - \{X\}} \R(Y)) \rightarrow \R(X)$.
This mathematical notation just makes precise the fact that 
$F_X$ determines the value of $X$,
given the values of all the other variables in $\U \union \V$.
If there is one exogenous variable $U$ and three endogenous
variables, $X$, $Y$, and $Z$, then $F_X$ defines the values of $X$ in
terms of the values of $Y$, $Z$, and $U$.  For example, we might have 
$F_X(u,y,z) = u+y$, which is usually written as
$X\gets U+Y$.%
\footnote{The fact that $X$ is assigned  $U+Y$ (i.e., the value
of $X$ is the sum of the values of $U$ and $Y$) does not imply
that $Y$ is assigned $X-U$; that is, $F_Y(U,X,Z) = X-U$ does not
necessarily hold.}  Thus, if $Y = 3$ and $U = 2$, then
$X=5$, regardless of how $Z$ is set.  

In the running forest-fire example, suppose that we have an exogenous
random variable $U$ that determines the values of $L$ and $\ML$.  Thus, $U$ has
four possible values of the form $(i,j)$, where both of $i$ and $j$ are
either 0 or 1.  The $i$ value determines the value of $L$ and the $j$
value determines the value of $\ML$.  Although $F_L$ gets as arguments
the vale of $U$, $\ML$, and $\FF$, in fact, it depends only
on the (first component of) the value of $U$; that is,
$F_L((i,j),m,f) = i$.  Similarly, $F_{\ML}((i,j),l,f) = j$.
The value of $\FF$ depends only on the value of $L$ and $\ML$.
\emph{How} it depends on them depends on whether 
either cause by itself is sufficient for the forest fire
or whether both are
necessary.  If either one suffices, then $F_{\FF}((i,j),l,m) =
\max(l,m)$,
or, perhaps more comprehensibly, $\FF =\max(L,\ML)$; if both are needed,
then $\FF = \min(L,\ML)$.  For future reference, call the former model
the \emph{disjunctive} model, and the latter the \emph{conjunctive} model.

The key role of the structural equations is to define what happens in the
presence of external interventions.  For example, we can explain what
happens if the arsonist does \emph{not} drop the match.
In the disjunctive model, there is a forest fire 
exactly if there is lightning; in the conjunctive model, there is
definitely no fire.
Setting the value of some variable $X$ to $x$ in a causal
model $M = (\S,\F)$ results in a new causal model denoted $M_{X
\gets x}$.  In the new causal model, 
the equation for $X$ is very
simple: $X$ is just set to $x$; the remaining equations are unchanged.
More formally, 
$M_{X \gets x} = (\S, \F^{X \gets x})$, where 
$\F^{X \gets x}$ is the result of replacing the equation for $X$ in $\F$
by $X = x$.

The structural equations describe \emph{objective}
information about the results of interventions, that can, in principle,
be checked.  Once the modeler has selected a set of variables to include in the model,
\emph{the world} determines which equations among those variables correctly represent the 
effects of interventions.\footnote{In general, there may be uncertainty about the causal model,
as well as about the true setting of the exogenous variables in a causal
model.  Thus, we may be uncertain about whether smoking causes cancer
(this represents uncertainty about the causal model) and uncertain about
whether a particular patient actually smoked (this is uncertainty about
the value of the exogenous variable that determines whether the patient
smokes).  This uncertainty can be described by putting a probability on
causal models and on the values of the exogenous variables.  We can then
talk about the probability that $A$ is a cause of $B$.} 
By contrast, the \emph{choice} of variables is
subjective; in general, there need be no objectively  ``right'' set of exogenous and 
endogenous variables to use in modeling a problem.    We return to this issue in
Section~\ref{sec:variables}.

It may seem somewhat circular to use
causal models, which clearly already encode causal information, to
define actual causation.  Nevertheless, as we shall see, there is no
circularity.  The equations of a causal model do not represent relations of \emph{actual causation}, 
the very concept that we are using them to define. Rather, the equations characterize 
the results of \emph{all possible} interventions (or at any rate, all of the interventions 
that can be represented in the model) without regard to what actually happened. 
Specifically, the equations do not depend upon the actual values realized by the 
variables.  For example, the equation $\FF = \max(L,\ML)$, by itself, does not say anything about whether the forest fire was actually caused by lightning or by an arsonist, or, for that matter, whether a fire even occurred. By contrast, relations of actual causation depend crucially on how things 
actually play out.

A sequence of endogenous $X_1, \ldots, X_n$ of is a \emph{directed path}
from $X_1$ to $X_n$ if
the value of $X_{i+1}$ (as given by 
$F_{X_{i+1}}$) depends on the value of $X_i$, for $1 = 1, \ldots, n-1$.
In this paper, following HP, we restrict 
our discussion
to \emph{acyclic} causal
models, where causal influence can be represented by an acyclic Bayesian
network.  That is, there is no cycle $X_1, \ldots, X_n, X_1$ of
endogenous variables that forms a directed path from $X_1$ to itself.
If $M$ is an
acyclic  causal model, then given a \emph{context}, that is, a setting
$\vec{u}$ for the exogenous variables in $\U$, there is a unique
solution for all the equations.

\section{The HP Definition of Actual Cause}\label{sec:actcaus}

\subsection{A language for describing causes}

Given a signature $\S = (\U,\V,\R)$, a \emph{primitive event} is a
formula of the form $X = x$, for  $X \in \V$ and $x \in \R(X)$.  
A {\em causal formula (over $\S$)\/} is one of the form
$[Y_1 \gets y_1, \ldots, Y_k \gets y_k] \phi$,
where
$\phi$ is a Boolean
combination of primitive events,
$Y_1, \ldots, Y_k$ are distinct variables in $\V$, and
$y_i \in \R(Y_i)$.
Such a formula is
abbreviated
as $[\vec{Y} \gets \vec{y}]\phi$.
The special
case where $k=0$
is abbreviated as
$\phi$.
Intuitively,
$[Y_1 \gets y_1, \ldots, Y_k \gets y_k] \phi$ says that
$\phi$ would hold if
$Y_i$ were set to $y_i$, for $i = 1,\ldots,k$.

A causal formula $\psi$ is true or false in a causal model, given a
context.
As usual, we write $(M,\vec{u}) \sat \psi$  if
the causal formula $\psi$ is true in
causal model $M$ given context $\vec{u}$.
The $\sat$ relation is defined inductively.
$(M,\vec{u}) \sat X = x$ if
the variable $X$ has value $x$
in the
unique (since we are dealing with acyclic models) solution
to
the equations in
$M$ in context $\vec{u}$
(that is, the
unique vector
of values for the endogenous variables that simultaneously satisfies all
equations 
in $M$ 
with the variables in $\U$ set to $\vec{u}$).
The truth of conjunctions and negations is defined in the standard way.
Finally, 
$(M,\vec{u}) \sat [\vec{Y} \gets \vec{y}]\phi$ if 
$(M_{\vec{Y} \gets \vec{y}},\vec{u}) \sat \phi$.
We write $M \sat \phi$ if $(M,\vec{u}) \sat \phi$ for all contexts $\vec{u}$.

For example, if $M$ is the disjunctive causal model for the forest
fire, and $u$ is the context where there is lightning and the arsonist
drops the lit match, 
then $(M,u) \sat \mbox{$[\ML \gets 0](\FF=1)$}$, 
since even if the arsonist is somehow
prevented from dropping the match, the forest burns (thanks to the
lightning);  similarly, $(M,u) \sat [L \gets 0](\FF=1)$.  However,
$(M,u) \sat [L \gets 0;\ML \gets 0](\FF=0)$: if the arsonist does not drop
the lit match 
and the lightning does not strike, then the forest does not burn.

\subsection{A preliminary definition of causality}
The HP definition of causality, like many others, is based on
counterfactuals.  The idea is that 
if $A$ and $B$ both occur, then $A$ is a cause of $B$ if, if $A$
hadn't occurred, then $B$ would not have occurred.
This idea goes back to at least Hume \citeyear[Section
{VIII}]{hume:1748}, who said:
\begin{quote}
We may define a cause to  be an object followed
by another, \ldots, where, if the first object had not been, the second
never had existed.
\end{quote}
This is essentially the \emph{but-for} test, perhaps the most widely
used test of actual causation in tort adjudication.  The but-for test
states that an act is a cause of injury if and only if, but for the act
(i.e., had the the act not occurred), the injury would not have
occurred.

There are two well-known problems with this definition.  The first can
be seen by considering the disjunctive causal model for the forest fire
again.  Suppose that the 
arsonist drops a match and lightning strikes.
Which is the cause?
According to a naive interpretation of the counterfactual definition,
neither is.  If the match hadn't dropped, then the lightning would
still have struck, so there would have been a forest fire anyway.
Similarly, if the lightning had not occurred, there still would have
been a forest fire.  As we shall see, the HP definition declares
both lightning and the 
arsonist 
causes
of the fire.  (In general, there may be more than one
actual cause of an outcome.)

A more subtle problem is what philosophers have called
\emph{preemption}, which is illustrated by the rock-throwing example
from the introduction.   As we observed, according to a naive
counterfactual definition of causality, Suzy's throw would not be a
cause.  

The HP definition deals with the first  problem by defining causality as 
counterfactual dependency \emph{under certain contingencies}.
In the forest-fire example, the forest fire does counterfactually depend
on the lightning under the contingency that the arsonist does not drop
the match; similarly, the forest fire depends counterfactually on the
dropping of the match under the contingency that the lightning does not
strike.  

Unfortunately, we cannot use this simple solution to treat the case of
preemption.   
We do not want to make Billy's
throw the cause of the bottle shattering by considering the contingency
that Suzy does not throw.   So if our account is to yield the correct
verdict in this case,  
it will be necessary to limit the contingencies that can be considered.
The reason that we consider Suzy's throw to
be the cause and Billy's throw not to be the cause is that Suzy's rock
hit the bottle, while Billy's did not.  Somehow the definition of actual
cause must 
capture this obvious intuition.

With this background, we now give the preliminary version of the HP definition
of causality.  Although the definition is labeled ``preliminary'',
it is quite close to the final definition, which is given 
in Section~\ref{sec:final}.  The definition is relative to a causal
model (and a context); $A$ may be a cause  of $B$ in one causal model
but not in another. The definition consists of three clauses.  The first
and third are quite simple; all the work is going on in the second clause.  

The types of events that the HP definition allows as actual causes are
ones of the form $X_1 = x_1 \land \ldots \land X_k = x_k$---that is,
conjunctions of primitive events; this is often abbreviated as $\vec{X}
= \vec{x}$. The events that can be caused are arbitrary Boolean
combinations of primitive events.
The definition does not allow statements of the form  ``$A$ or $A'$ is a
cause of $B$'', although this could be treated as being equivalent to
``either $A$ is a cause of $B$ or $A'$ is a cause of $B$''.    
On the other hand, statements such as
``$A$ is a cause of $B$ or $B'$'' are allowed;  this
is not equivalent to ``either $A$ is a cause of $B$ or $A$ is a cause
of $B'$''.

\dfn\label{actcaus}
(Actual cause; preliminary version) \cite{HP01b}
$\vec{X} = \vec{x}$ is an {\em actual cause of $\phi$ in
$(M, \vec{u})$ \/} if the following
three conditions hold:
\begin{description}
\item[{\rm AC1.}]\label{ac1} $(M,\vec{u}) \sat (\vec{X} = \vec{x})$ and 
$(M,\vec{u}) \sat \phi$.
\item[{\rm AC2.}]\label{ac2}
There is a partition of $\V$ (the set of endogenous variables) into two
subsets $\vec{Z}$ and $\vec{W}$  
with $\vec{X} \subseteq \vec{Z}$, and a
setting $\vec{x}'$ and $\vec{w}$ of the variables in $\vec{X}$ and
$\vec{W}$, respectively, such that
if $(M,\vec{u}) \sat Z = z^*$ for 
all $Z \in \vec{Z}$, then
both of the following conditions hold:
\begin{description}
\item[{\rm (a)}]
$(M,\vec{u}) \sat [\vec{X} \gets \vec{x}',
\vec{W} \gets \vec{w}]\neg \phi$.
\item[{\rm (b)}]
$(M,\vec{u}) \sat [\vec{X} \gets
\vec{x}, \vec{W}' \gets \vec{w}, \vec{Z}' \gets \vec{z}^*]\phi$ for 
all subsets $\vec{W}'$ of $\vec{W}$ and all subsets $\vec{Z'}$ of
$\vec{Z}$, where we abuse notation and write $\vec{W}' \gets \vec{w}$ to
denote the assignment where the variables in $\vec{W}'$ get the same
values as they would in the assignment $\vec{W} \gets \vec{w}$. 
\end{description}
\item[{\rm AC3.}] \label{ac3}
$\vec{X}$ is minimal; no subset of $\vec{X}$ satisfies
conditions AC1 and AC2.
\label{def3.1}  
\end{description}
\edfn

AC1 just says that $\vec{X}=\vec{x}$ cannot
be considered a cause of $\phi$ unless both $\vec{X} = \vec{x}$ and
$\phi$ actually happen.  AC3 is a minimality condition, which ensures
that only those elements of 
the conjunction $\vec{X}=\vec{x}$ that are essential for
changing $\phi$ in AC2(a) are
considered part of a cause; inessential elements are pruned.
Without AC3, if dropping a lit match qualified as a
cause of the forest fire, then dropping a match and
sneezing would also pass the tests of AC1 and AC2.
AC3 serves here to strip ``sneezing''
and other irrelevant, over-specific details
from the cause.
Clearly, all the ``action'' in the definition occurs in AC2.
We can think of the variables in $\vec{Z}$ as making up the ``causal
path'' from $\vec{X}$ to $\phi$, 
consisting of one or more directed paths from variables in 
$\vec{X}$ to variables in $\phi$.  
Intuitively, changing the 
value(s)
of
some 
variable(s)
in 
$\vec{X}$
results in changing the value(s) of some
variable(s) in $\vec{Z}$, which results in the 
value(s)
of some
other variable(s) in $\vec{Z}$ being changed, which finally results in
the truth value of $\phi$ changing.  The remaining endogenous variables, the
ones in $\vec{W}$, are off to the side, so to speak, but may still have
an indirect effect on what happens.
AC2(a) is essentially the standard
counterfactual definition of causality, but with a twist.  If we 
want to show that $\vec{X} = \vec{x}$ is a cause of $\phi$, we must show
(in part) that if $\vec{X}$ had a different value, then 
$\phi$ would have been false.  However, this effect of the value of
$\vec{X}$ on the  
truth value of
$\phi$ may not hold in the actual context;
the value of $\vec{W}$ may have to be different to allow this
effect to manifest itself.  For example, consider 
the context where both the lightning
strikes and the arsonist drops a match in the disjunctive model of the
forest fire.  Stopping the arsonist from
dropping the match will not prevent the forest fire.  The
counterfactual effect of the arsonist on the forest fire manifests
itself only in a situation where the lightning does not strike (i.e., where
$L$ is set to 0).  AC2(a) is what allows us to call both the
lightning and the arsonist causes of the forest fire.
Essentially, it ensures that
$\vec{X}$ alone suffices to bring about the change from $\phi$ to $\neg
\phi$; setting $\vec{W}$ to $\vec{w}$ merely eliminates
possibly spurious side effects that may mask the effect of changing the
value of $\vec{X}$.  Moreover, 
when $\vec{X} = \vec{x}$,
although the values of variables on the
causal path (i.e.,  the variables $\vec{Z}$) may be perturbed by
the change to $\vec{W}$, this perturbation has no impact on the value of
$\phi$.  If  $(M,\vec{u}) \sat \vec{Z} = \vec{z}^*$, then 
$z^*$
is the value of the 
variable $Z$ in the context $\vec{u}$.  We capture the fact that the
perturbation has no impact on the value of $\phi$ by saying that if some
variables $Z$ on the causal path were set to their original values in the
context $\vec{u}$, $\phi$ would still be true, as long as $\vec{X} =
\vec{x}$.

\xam\label{ex:ff}
For the forest-fire example, 
let  $M$ be the 
disjunctive model for the forest fire sketched earlier, 
with endogenous variables $L$, $\ML$, and $\FF$.
We want to show that $L=1$ is an actual cause of $\FF=1$.
Clearly 
$(M,(1,1)) \sat \FF=1$ and 
$(M,(1,1)) \sat L=1$; in the context (1,1), the lightning strikes
and the forest burns down.  Thus, AC1 is satisfied.  AC3 is trivially
satisfied, since $\vec{X}$ consists of only one element, $L$, so must be
minimal.  For AC2, take $\vec{Z} = \{L, \FF\}$ and take $\vec{W} =
\{\ML\}$, let $x' = 0$, and let $w = 0$.  Clearly,
$(M,(1,1)) \sat [L \gets 0, \ML \gets 0](\FF \ne 1)$; if the lightning
does not strike and the match is not dropped, the forest does not burn
down, so AC2(a) is satisfied.  To see the effect of the lightning, we
must consider the contingency where the match is not dropped; the
definition allows us to do that by setting $\ML$ to 0.  (Note that here
setting $L$ and $\ML$ to 0 overrides the effects of $U$; this is
critical.)  Moreover,  
$(M,(1,1)) \sat [L \gets 1, \ML \gets 0](\FF = 1)$;
if the lightning
strikes, then the forest burns down even if the lit match is not
dropped, so AC2(b) is satisfied.  (Note that since $\vec{Z} = \{L, \FF\}$, 
the only subsets of $\vec{Z} - \vec{X}$ are the empty set and the
singleton set consisting of just $\FF$.)

It is also straightforward to show that 
the lightning and the dropped match are also causes of the forest fire
in the context where $U = (1,1)$ in the conjunctive model.
Again, AC1 and AC3 are trivially satisfied and, again, to show that 
AC2 holds in the case of lightning we can
take $\vec{Z} = \{L, \FF\}$, $\vec{W} =
\{\ML\}$, and $x' = 0$, but now we let  $w = 1$.
In the conjunctive scenario, if there is no lightning, there is no
forest fire, while if there is lightning (and the match is dropped)
there is a forest fire,
so AC2(a) and AC2(b) are satisfied; similarly 
for the dropped match.
\exam

\xam\label{xam2}
Now consider the Suzy-Billy example.%
\footnote{The discussion of this example is taken
almost verbatim from HP.}
We get the desired result---that Suzy's throw is a cause, but Billy's is
not---but only if we 
model the story appropriately.  Consider first a coarse causal
model, with three endogenous variables:
\begin{itemize}
\item $\ST$ for ``Suzy throws'', with values 0 (Suzy does not throw) and
1 (she does);
\item $\BT$ for ``Billy throws'', with values 0 (he doesn't) and
1 (he does);
\item $\BS$ for ``bottle shatters'', with values 0 (it doesn't shatter)
and 1 (it does).
\end{itemize}
(We omit the exogenous variable here; it determines whether Billy and
Suzy throw.)  Take the formula for $\BS$ to be such that the bottle
shatters if either Billy or Suzy throw; that is 
$\BS = \max(\BT, \ST)$.
(We assume that Suzy and Billy will not miss if they throw.)
$\BT$ and $\ST$ play symmetric roles in this model;
there is nothing to distinguish them.
Not surprisingly, both
Billy's throw and Suzy's throw are classified as causes of the
bottle shattering
in this model.  
The argument is essentially identical to that in the disjunctive model
of the forest-fire example in the context 
$U = (1,1)$, where both the lightning and 
the dropped match are causes of the fire.

The trouble with this model is that it cannot distinguish
the case where both rocks
hit the bottle simultaneously (in which case it would be reasonable
to say that both $\ST=1$ and $\BT=1$ are 
causes of $\BS=1$) from the case where Suzy's rock
hits first.  
\commentout{The model has to be refined to express this distinction.
One way is to invoke a dynamic model \cite[p.~326]{pearl:2k}.  This
model is discussed by HP; we omit details here.
A perhaps simpler way to gain expressiveness is to allow $\BS$ to be
three valued, with values 0 (the bottle doesn't shatter), 1
(it shatters as a result of being hit by Suzy's rock), and 2 (it
shatters as a result of being hit by Billy's rock).
We leave it to the reader to check that $\ST = 1$ is a
cause of $\BS = 1$, but $\BT = 1$ is not (if Suzy doesn't thrown but
Billy does, then we would have $\BS = 2$).  Thus, to some extent, this
solves our problem.  But it
borders on cheating; the answer is almost
programmed into the model by invoking the relation ``as a result of'',
which requires the identification of the actual cause.
}
To allow the model to express this distinction,
we add two new variables to the model:
\begin{itemize}
\item $\BH$ for ``Billy's rock hits the (intact) bottle'', with values 0
(it doesn't) and 1 (it does); and
\item $\SH$ for ``Suzy's rock hits the bottle'', again with values 0 and
1.
\end{itemize}
Now our equations will include:
\begin{itemize}
\item $\SH = \ST$;
\item $\BH = \min(\BT, 1-\SH)$; and
\item $\BS = \max(\SH, \BH)$.
\end{itemize}
Now it is the case that, in the context where both Billy and Suzy throw,
$\ST=1$ is a cause 
of $\BS=1$, but $\BT = 1$ is not.
To see that $\ST=1$ is a cause, note that, as usual, it is immediate
that AC1 and AC3 hold.  For AC2, choose 
$\vec{Z} = \{\ST,\SH,\BH,\BS\}$,
$\vec{W}=\{\BT\}$,  and $w=0$. 
When $\BT$ is set to 0, $\BS$ tracks $\ST$: if Suzy
throws, the bottle shatters  and if she doesn't throw, the bottle does
not shatter.  To see that $\BT=1$ is \emph{not} a cause
of $\BS=1$, we must check that there is no
partition $\vec{Z} \cup \vec{W}$ of the endogenous variables that
satisfies AC2.
Attempting the symmetric choice with 
$\vec{Z} = \{\BT,\BH,\SH,\BS\}$,
$\vec{W}=\{\ST\}$, and $w=0$ 
violates AC2(b). To see this, take $\vec{Z}' = \{\BH\}$.  In the
context where Suzy 
and Billy both throw, $\BH=0$.  If $\BH$ is set to 0, the bottle does
not shatter if Billy throws and Suzy does not.  
It is precisely because, in this context, Suzy's throw hits the bottle
and Billy's does not that we declare Suzy's throw to be the cause of the
bottle shattering.  AC2(b) captures that intuition by allowing us to
consider the contingency where $\BH=0$, despite the fact that Billy
throws.   We leave it to the reader to check that no other partition of
the endogenous variables satisfies AC2 either.  


This example emphasizes an important moral.
If we want to argue in a case of preemption
that $X=x$ is the cause of $\phi$ rather than $Y=y$,
then there must be a random variable ($\BH$ in this case) that takes on
different values depending on whether $X=x$ or $Y=y$ is the actual
cause.  If the model does not contain such a variable, then it will not
be possible to determine which one is in fact the cause.  This is
certainly consistent with intuition and the way we present evidence.  If
we want to  argue (say, in a court of law) that it was $A$'s shot that
killed $C$ rather than $B$'s, then we present evidence such as the
bullet entering $C$ from the left side (rather than the right side, which
is how it would have entered had $B$'s shot been the lethal one).
The side from which the shot entered is the relevant random variable in
this case.  Note that the random variable may involve temporal evidence
(if $Y$'s shot had been the lethal one, the death would have occurred
a few seconds later), but it certainly does not have to.
\exam


%

\commentout{
The structural equations in a causal model describe objective features
of the world. Given a set of variables Ð or at any rate, given a ``good"
set of variables, free of the sorts of defects discussed below Ð ``the
world" determines which equations hold among those variables. The
equations of a causal model thus constitute the objective core of the
model. It is these equations, rather than the relations of actual
causation entailed by a model, that are beholden to the world. In this
way, we can allow that relations of actual causation are partly a
function of how we choose to model the world, without denying the
objectivity of causal relations. 

A causal model performs a number of different tasks.  It makes pronouncements about actual causation, but in addition, it makes empirical predictions about the relationships that will hold among the values of the variables, both under observation, and under intervention. If we observe a causal system without intervening, the model tells us that we will observe values of the variables that satisfy the equations of the model. If we intervene on one or more of the variables, the model tells us that the values of the variables that result will satisfy the equations of the appropriately modified model. 

These predictions provide independent constraints upon the equations that can be included in a satisfactory causal model. If the only constraints were that the model should be correct in its pronouncements concerning actual causation, then the resulting flexibility on the part of the model-builder would render the HP account of actual causation vacuous. The modeler could simply adjust the equations in whatever way necessary to yield the desired relations of actual causation. But because the very same model must also yield predictions about observations and interventions, the modeler's flexibility is limited considerably. 

[discuss example from earlier illustration]
}

\section{The Choice of Variables}\label{sec:variables}

A modeler has considerable leeway in choosing which variables to include
in a model.  Nature does not provide a uniquely correct set of
variables. Nonetheless, there 
are a number of considerations that guide variable selection. While
these will not usually suffice to single out one choice of variables,
they can provide a framework for the rational evaluation of models,
including resources for motivating and defending certain choices of
variables, and criticizing others.  

The problem of choosing a set of variables for inclusion in a model has
many dimensions. One set of issues concerns the question of how many
variables to include in a model. If the modeler begins with a set of
variables, how can she know whether she should add additional variables
to the model? Given that it is always possible to add additional
variables, is there a point at which the model contains ``enough"
variables? Is it ever possible for a model to have ``too many"
variables? 
Can the addition of further variables ever do positive harm to a model?  



\commentout{ 
Another set of issues concerns the possible values of a given variable
in a model. In probability theory, one begins with an outcome space, and
a random variable corresponds to a partition of the outcome space. But
in a causal model, the outcome space is not antecedently given: the
range of values of a variable partially determine the set of
possibilities allowed by the model. Are there conditions in which the
set of possibilities should be expanded or contracted? That is, if we
have a model that includes a variable $X$ with a particular range, are
there considerations that would lead us to prefer a model that replaced
$X$ with a new variable $X'$, whose range is a proper subset or superset
of the range of $X$?  
}

Another set of issues concerns the values of variables.
Say that variable $X'$ is
a \emph{refinement} of $X$ if, for each value $x$ in the range of $X$,
there is 
some subset  
$S$
of the range of $X'$ such that $X = x$ just in case $X'$ is in 
$S$. 
When is it appropriate or desirable to replace a variable with a
refinement?
Can it ever lead to problems if a variable is too fine-grained? Similarly, are
there considerations that would lead us to prefer a model that replaced
$X$ with a new variable $X''$, whose range is a proper subset or superset
of the range of $X$?

Finally, are there constraints on the set of variables in a model over and above
those we might impose on individual variables? For instance, can the
choice to include a particular variable $X$ within a model require us to
include another variable $Y$, or to exclude a particular variable $Z$? 

\commentout{
It would not be surprising if these considerations interact with one
another. For instance, it may well be that the choice of whether to
include a more coarsely-grained variable $X$, or a refinement  
$X'$, will determine whether another variable $Y$ ought to be included
or excluded. Moreover, it is to be expected that considerations of which
variables to include in a model will interact with the sorts of
constraints upon the equations of a causal model discussed above. 
}

While we cannot provide complete answers to all of these questions,
we believe a good deal can be said to reduce the arbitrariness of the
choice of variables.
The most plausible way to motivate guidelines for the selection of
variables is to show how inappropriate choices 
give rise to systems of equations that are inaccurate,
misleading, or incomplete in their predictions of observations and
interventions.  
In the next three subsections, we present several examples to
show how such considerations can be brought to bear
on the problem of variable choice.

\subsection{The Number of Variables}


\commentout{
The values of a variable in a model correspond to possibilities that can
be contemplated within the model. Variables are rarely total functions,
in the sense that their values span the entire space of logical
possibilities. As such, most variables embody certain presuppositions
that get held fixed within the model. For example, a model might contain
a variable $X$, where $X$ = 1 represents Suzy's throwing a stone, and
$X$ = 0 represents her failing to throw a stone. This variable has as a
presupposition that Suzy exists Ð there is no value of the variable that
corresponds to Suzy's non-existence. Thus the inclusion of such a
variable in the model builds Suzy's existence into the model as a
presupposition.  
}

We already saw in Example~\ref{xam2} that it is important to choose the
variables correctly.  Adding more variables can clearly affect whether
$A$ is a cause of $B$.
\commentout{
[[SHOULD WE SAY SOMETHING ABOUT STABILITY HERE?  DO WE HAVE ANYTHING TO
SAY ABOUT WHEN WE HAVE ``ENOUGH'' VARIABLES.]]
}
When is it appropriate or necessary to add further variables to a model?%
\footnote{Although his model of causality is quite different from ours, 
Spohn \citeyear{Spohn03} also considers the effect of adding or removing
variables, and discusses 
how a model with fewer variables should be related to one with more
variables.} 
Suppose that we have an infinite sequence of models $M^1, M^2, \ldots$
such that the variables in $M^{i}$ are $X_0, \ldots, X_{i+1},Y$, and 
$M^{i+1}_{X_{i+1} \gets 1} = M_i$ (so that $M^{i+1}$ can be viewed as an
extension of $M^i$).  Is it possible that whether $X_0 = 1$ is a cause
of $Y=1$ can alternate as we go through this sequence?  This would
indicate a certain ``instability'' in the causality.  
In this circumstance, a lawyer
should certainly be able to argue against using, say, $M^7$ as a model to show that $X_0
= 1$ is  cause of $Y=1$. 
On the other hand, if the sequence stabilizes, that is, if there is some $k$ such that 
for all $i \ge k, M^i$ delivers the same verdict on some causal claim of interest, 
that would provide a strong reason to accept $M^k$ as sufficient. 

Compare Example~\ref{ex:ff} with Example~\ref{xam2}. 
In Example~\ref{ex:ff}, we were able to 
adequately model the scenario using only three endogenous 
variables: $L$, $\ML$, and $\FF$.
By contrast, in Example~\ref{xam2}, the model containing only
three endogenous variables, $\BT$, $\ST$, and $\BS$, was inadequate.
What is the difference between the two scenarios?
One difference we have already mentioned is that there seems to be an
important feature of the second scenario that cannot be captured in the 
three-variable model: Suzy's rock hit the bottle before Billy's did.
There is also a significant ``topological'' difference between the two
scenarios. In 
the forest-fire example, there are two directed paths into the variable $\FF$.
We could interpolate additional variables along these two paths. We could, 
for instance, interpolate a variable representing the occurrence of a
small brush fire. But doing so would not fundamentally change
the causal structure: there would still be just two directed paths into $\FF$.
In the case of preemption, however, adding the additional variables $\SH$
and $\BH$ created an additional directed path that was not there before.
The three-variable model contained just two directed paths: one from $\ST$ 
to $\BS$, and one from $\BT$ to $\BS$. However, once the variables 
$\SH$ and $\BH$ were added, there were three directed paths: $\{\ST, \SH, \BS\}$, 
$\{\BT, \BH, \BS\}$, and $\{\ST, \SH, \BH, \BS\}$. 
The intuition, then, is that adding additional variables to a model
will not 
affect the relations of actual causation that hold in
the model unless the addition of those variables changes the
``topology" of the model.
\commentout
{
We may make this intuition precise in the form of a conjecture. 
Let $\vec{P} = \{X_1, \ldots, X_n\}$ be a directed path in a causal model $M$.
We say that the variable $X_i$ is \emph{independent} of  $\vec{P}$ just in
case $X_i$ 
does not lie on any directed path that is not a sub-path or super-path
of $\vec{P}$.  
Then we conjecture that if the addition of variable to a causal model does not
either create a new directed path, or separate a directed path that was
previously unseparated, then the addition of that variable will not change any relations
of actual causation.
}
\commentout
{
Suppose that we have an infinite sequence of models $M^1, M^2, \ldots$
such that the variables in $M^{i}$ are $X_0, \ldots, X_{i+1},Y$, and 
$M^{i+1}_{X_{i+1} \gets 1} = M_i$ (so that $M^{i+1}$ can be viewed as an
extension of $M^i$.)  Is it possible that whether $X_0 = 1$ is a cause
of $Y=1$ can alternate as we go through this sequence?  This would
indicate a certain ``instability'' in the causality.  Certainly a lawyer
should be able to argue against using $M^7$ as a model to show that $X_0
= 1$ is  cause of $Y=1$. 
} 
A more complete mathematical characterization of the conditions under
which the verdicts of actual causality remain stable under the addition
of further 
variables strikes us as a worthwhile research project that has not yet
been undertaken.

\subsection{The Ranges of Variables}

Not surprisingly, the set of possible values of a variable must also be
chosen appropriately.
Consider, for example, a case of
``trumping'', introduced by Schaffer \citeyear{Schaffer01}. Suppose that 
a group of soldiers is very well trained, so that they will obey any
order given by 
a
superior officer; in the case of conflicting
orders, they obey the highest-ranking officer. Both a sergeant and a
major issue the order 
to march, and the soldiers march. 
Let us put aside the morals that
Schaffer attempts to draw from this example (with which we disagree; 
see \cite{HP01b} and \cite{Hitchcock09}), and consider only the modeling
problem.  We will presumably want variables $S$, $M$, and
$A$, corresponding to the sergeant's order, the major's order, and the
soldiers' action. We might let $S$ = 1 represent the sergeant's giving
the order to march and $S$ = 0 represent the sergeant's giving no order;
likewise for $M$ and $A$. 
%
%
%
%
But this would not be adequate.  If the only possible order is the order
to march, then there is no way to capture the principle that in the case
of conflicting orders, the soldiers obey the major.  One way to do this
is to replace the variables $M$, $S$, and $A$ by variables $M'$, $S'$ and $A'$ that
take on three possible values.  Like $M$, $M'=0$ if the major gives no
order and $M'=1$ if the major gives the order to march.  But now we
allow $M' = 2$, which corresponds to the major giving some other order.
$S'$ and $A'$ are defined similarly.  We can now write an equation to capture the
fact that if $M'=1$ and $S'=2$, then the soldiers march, while if $M'=2$
and $S'=1$, then the soldiers do not march.

%
%

The appropriate set of values of a variable will depend on the other
variables in the picture, and the relationship between them.  Suppose, for
example, that a hapless homeowner comes home from a trip to find that
his front door is stuck. If he pushes on it with a normal force
then the door will not open. However, if he leans
his shoulder against it and gives a solid push, then the door will
open.  To model this, it suffices to have a variable $O$ with values
either 0 or 1, depending on whether the door opens, and a variable $P$,
with values 0 or 1 depending on whether or not the homeowner gives a
solid push.  

%

%
%
%
On the other hand, suppose that the homeowner also forgot to disarm the
security system, and that the system is very sensitive, so that it will
be tripped by any push on the door, regardless of whether the door
opens. Let $A$ = 1 if the alarm goes off, $A$ = 0 otherwise. 
Now if we try to model the situation with the same variable $P$,
we will not be able to express the dependence of the alarm
on the homeowner's push.
To deal with both $O$ and $A$, we need to extend $P$ to a 3-valued
variable $P'$, with values 0 if the homeowner does not push the door, 1
if he pushes it with normal force, and 2 if he gives it a solid push.

\commentout{
Note that we do \emph{not} require that a variable be in some sense
``maximally'' finely-grained.  For example, we do not require that a
variable that takes on all the possible degrees of force that the
homeowner could apply to the door; we just require that 
variables be sufficiently finely-grained to capture relevant
distinctions in a story.  A reasonable argument against a model that a
lawyer could present would be that the model is not able to capture
relevant distinctions, either because there are too few variables or
because the variables used do not have enough possible values.
}

These considerations parallel issues that arise in philosophical
discussions about the metaphysics of ``events".%
\footnote{
This philosophical usage of the word ``event"
is different from the typical usage of the word in computer
science and probability, where an event is just a subset of the state space.}
Suppose that our
homeowner pushed on the door with 
enough force to open it.
Is there just one event, the push, that can be
described at various levels of detail,
such as %
a ``push" or a ``hard push"?
This is the view of Davidson \citeyear{Davidson67}. Or are
there rather many different events corresponding to these different
descriptions,
as argued by Kim \citeyear{kim73} and Lewis \citeyear{Lewis86c}?
And if we take the latter
view, which of the many events that occur should be counted as causes of
the door's opening? 
These strike us as
pseudoproblems. 
We believe that questions about causality are best addressed by dealing
with the methodological problem of constructing a model that 
%
correctly describes the effects of interventions
in a way that is not misleading or ambiguous.

A slightly different way in which one variable may constrain 
the values that another may take is by its implicit presuppositions.
For example, a counterfactual theory of causation 
seems to have 
the somewhat counterintuitive
consequence that one's birth is a cause of one's death. This sounds a
little odd. If Jones dies suddenly one night, shortly before his 80th
birthday, the coroner's inquest is unlikely to list ``birth" as among
the causes of his death. 
Typically, when we investigate the causes of death,
we are interested in what makes the difference between
a person's dying and 
his
surviving.
So our model might include a variable $D$ such 
$D=1$ holds if Jones dies shortly before his 80th birthday, and $D=0$ holds if
he continues to live.  
If our model also includes a variable $B$,
taking the value 1 if Jones is born, 0 otherwise, 
then there simply is no value that $D$ would take
if $B=0$.  Both
$D=0$ and $D=1$ implicitly presuppose that Jones was born (i.e., $B=1$).
Our conclusion is that 
if we have chosen to include a variable such as $D$
in our model, then we cannot
conclude that
Jones' birth is a cause of his death!  
\commentout{
More generally, 
including the
variable $D$ in our model amounts to building in the presupposition that
Jones exists and lives to be 79. Only variables whose values are
compatible with that presupposition can be included in the model. This
may seem like an undue restriction, but 
we would argue that 
it in fact reflects an important
feature of the way we think about causation. 
}
	
\subsection{Dependence and Independence}

Lewis \citeyear{Lewis86a} added a constraint to his counterfactual theory of
causation. In order for event $c$ to be a cause of event $e$, the two
events cannot be logically related.
Suppose for
instance, that Martha says ``hello" loudly. If she had not said
``hello", then she certainly could not have said ``hello" loudly. But
her saying ``hello" is not a cause of her saying ``hello" loudly. The
counterfactual dependence results from a logical, rather than a causal,
relationship between the two events. 

We must impose a similar constraint upon causal models. 
Values of different variables should not correspond to events that are
logically related.
But now, rather than being an \emph{ad hoc} restriction, it has a clear
rationale.  For suppose that we had a model with variable $H_1$ and
$H_2$, where $H_1$ represents ``Martha says `hello'\,'' (i.e., $H_1 = 1$ if
Martha says ``hello'' and $H_1 = 0$ otherwise), and $H_2$ represents ``Martha
says `hello' loudly''.  The intervention $H_1 =0 \land H_2 = 1$ is
meaningless; it is logically impossible for Martha not to say 
``hello'' and to say `'hello'' loudly.


We doubt that any careful modeler would choose variables that have
logically related values.  
However, the converse of this principle, that
the different values of any particular variable \emph{should} be logically
related (in fact, mutually exclusive), is less obvious and equally important.
Consider Example~\ref{xam2}.
While, in the actual context, Billy's rock will hit the bottle just in case
Suzy's doesn't, this is not a necessary relationship.
Suppose that, instead of using two variables $\SH$ and $\BH$,
we try to
model the scenario with a variable 
$H$
that takes the value 1 if 
Suzy's rock hits, and 
and 0 if Billy's rock hits.
The reader can verify that, in this model, there is no contingency
such that the bottle's shattering depends upon Suzy's throw. 
The problem, as we said, is that $H=0$ and $H=1$ are \emph{not} mutually
exclusive;  
there are possible situations
in which both rocks hit or neither rock hits the bottle.
In particular, this representation does not allow us to consider
independent interventions on the rocks hitting the bottle. 
As the discussion in Example~\ref{xam2} shows, it is precisely such an
intervention that is needed to establish that Suzy's throw (and not
Billy's) is the actual cause of the bottle shattering.

While these rules are simple in principle, their application is not
always transparent. 
\xam\label{train}
Consider cases of ``switching", which have been much
discussed in the philosophical literature. A train is heading toward the
station. An engineer throws a switch, directing the train down the left
track, rather than the right track. The tracks re-converge
before the station, and the train arrives as scheduled. Was throwing the
switch a cause of the train's arrival? 
\commentout{
Let $A$ represent the arrival of
the train, and $S$ the throwing of the switch. At this point, we face a
choice about how to model the intermediate events. We might choose $L$
to represent the train's traveling on the left track, and $R$ to
represent its traveling on the right track. Then our equations will be:

\indent $S$ = $U$

\indent $L$ = $S$

\indent $R$ = 1 - $S$

\indent $A$ = $L$ or $R$

\noindent In this model, the HP account will rule that $S$ = 1 is an
actual cause of $A$ = 1. Letting \textbf{W} = \{$R$\}, and setting $R$ =
0, $A$ will depend upon $S$. On the other hand, if we instead use a
variable $T$, taking the value 2 if the train goes on the left hand
track, 1 if the train goes on the right hand track, and 0 if neither
(e.g. if it stops, or is de-railed), our equations will be:

\indent $S$ = $U$

\indent $T$ = $S$ + 1

\indent $A$ = 1 if $T$ > 0, $A$ = 0 otherwise.

\noindent (We may wish to add another variable for the presence of a
train, and make $T$ depend upon this variable -- we ignore this
complication.) Now the HP model will rule that $S$ = 1 is not an actual
cause of $A$ = 1. It's not immediately obvious which is the correct
model. The choice will depend upon what we take to be a conceivable
intervention. The first model permits independent interventions on the
two tracks; in particular, it permits interventions placing trains on
both tracks. If there really is only one train, and it cannot be in two
places at once, we may want to rule out the possibility of such an
intervention. 
}
HP consider two causal models of this scenario. 
In the first, there is a
random variable $S$ which is 1 if the switch is thrown (so the train
goes down the left track) and 0 otherwise.  In the second, 
in addition to $S$,
there are
variables $\LT$ and $\RT$, indicating whether or not the train goes down
the left track and right track, respectively.  Note that with the first
representation, there is no way to model the train not making it to the
arrival point.  With the second representation, we have the problem that
$\LT=1$ and $\RT=1$ are arguably not independent; the train cannot be on
both tracks at once.  If we want to model the possibility of one track
or another being blocked, we should use, instead of $\LT$ and $\RT$,
variables $\LB$ and $\RB$, which indicate whether the left track or
right track, respectively, are blocked.  This allows us to
represent all the relevant possibilities without running into
independence problems.  Note that if we have only $S$ as a random
variable, then $S=1$ cannot be a cause of the train arriving; it would
have arrived no matter what.  With $\RB$ in the picture, 
the preliminary HP definition of actual cause rules that
$S=1$ can be an
actual
cause of the train's arrival; for example, under the contingency that $\RB = 1$, the train
does not arrive if $S=0$.
(However, once we extend the definition to include defaults, as we will in the next section,
it becomes possible once again to block this conclusion.)
\exam

These rules will have particular consequences for how we should
represent events that might occur at different times. 
Consider the following simplification of an example introduced by
Bennett \citeyear{Bennett87}, and also considered in HP.
\xam
Suppose that the Careless Camper (CC for short) has plans to go
camping on the first weekend in June. He will go camping unless there is
a fire in the forest in May. If he goes camping, he will leave a
campfire unattended, and there will be a forest fire. 
Let the
variable $C$ take the value 1 if CC goes camping, and 0 otherwise. 
How should we represent the state of the forest?  

There appear to be at least three alternatives. The simplest proposal
would be to use a variable $F$ that takes the value 1 if there is a
forest fire at some time, and 0 otherwise.\footnote{This is, in effect,
how effects have been represented using ``neuron diagrams" in late
preemption cases. See Hitchcock \citeyear[pp. 85--88]{Hitchcock07b} for
discussion.} 
But now how are we to represent the dependency relations between $F$ and
$C$? Since CC will go camping only if there is no fire (in May), we
%
would want to have an equation such as $C = 1 - F$.
On the other hand, since there will be a fire (in June) just in case CC
goes camping, we will also need $F = C$.
This representation is clearly not rich enough, since it does not let us
make the clearly relevant distinction between whether the forest fire
occurs in May or June.  The problem is manifested in the fact that the
equations are cyclic, and have no consistent solution.%
\footnote{Careful readers will note the the preemption case of
Example~\ref{xam2} is modeled in this way. In that model,
$\BH$ is a cause of $\BS$, even though it is the earlier shattering
of the bottle that prevents Billy's rock from hitting. Halpern and Pearl
\citeyear{HP01b} note this problem and offer a dynamic model
akin to the one recommended below. As it turns out, this does
not affect the analysis of the example offered above.}

A second alternative, adopted by Halpern and Pearl
\citeyear[p.~860]{HP01b}, 
would be to use a variable $F'$ that takes the value
0 if there is no fire, 1 if there is a fire in May, and 2 if there is a
fire in June. 
\commentout{
While 
such a variable will certainly allow us to distinguish
between a fire in May and a fire in June,
the values $F=1$ and $F=2$ are not mutually exclusive; we cannot
represent the possibility of a fire in both May and June.
}
Now how should we write our
equations? Since CC will go camping unless there is a fire in May, the
equation for $C$ should say that $C = 0$ iff $F' = 1$.
And since there will be a fire in June if CC goes camping, the equation
for $F'$ should say that $F'$ = 2 if $C$ = 1 and $F'$ = 0 otherwise. 
These equations are cyclic.
Moreover, while they do have a consistent solution, 
they
are highly misleading in what they predict about the
effects of interventions. For example, the first equation tells us that
intervening to create a forest fire in June would cause CC to go camping
in the beginning of June. But this seems to get the causal order
backwards!
\commentout{
Certainly if we wanted CC to go camping (so we won't have to
put up with his loud parties that weekend), it would not be a very
effective strategy to plan to ignite a forest fire $after$ CC had
returned home. We might try to reason as follows: In order to ignite a
forest fire in June, we would have to ensure that the forest does not
burn in May. Thus our contemplated intervention has two parts: one part
to preserve the forest in May, and a second part to ignite it
June. True, the second part is redundant for purposes of sending CC on
his camping trip, but the strategy would nonetheless be effective. But
this line of reasoning departs from the usual way of thinking about
interventions. An intervention is a local, surgical manipulation of the
value of a variable that overrides the existing causal structure. For
example, in our model where Billy and Suzy are armed with stones, if we
intervene on whether or not the bottle shatters, we do not do so by
changing what Billy and Suzy do. Rather, an intervention involves an
independent causal process (a ``miracle" in the framework of Lewis 1979)
that shatters the bottle or protects it, independently of whatever it is
that Billy and Suzy do. Similarly, if we intervene to set fire to the
forest in June, we do not do so by protecting it in May, and allowing
the regular course of events produce a fire. Rather, we douse it in
gasoline (either literally or figuratively) and set it ablaze,
regardless of what has come before. And it should be apparent that this
sort of intervention, performed in June, could not influence CC's
earlier actions. 
A further problem with this representation
is that there is no consistent solution to the equations in which
F' = 1. 
In the actual context, there is no fire in May, so this is not a
problem. But ideally, the model should allow for alternative values of
the exogenous variables, representing, for example, the possibility of
lightning strikes in May, that could result in a May fire. The
impossibility of such solutions is a drawback of the model.  
}

The third way to model the scenario is to use two separate variables,
$F_{1}$ and $F_{2}$, to represent the state of the forest at separate
times. $F_{1}$ = 1 will represent a fire in May, and $F_{1}$ = 0
represents no fire in May; $F_{2}$ = 1 represents a fire in June and
$F_{2}$ = 0 represents no fire in June. Now we can write our equations
as $C = 1 - F_{1}$ 
and $F_{2} = C \times
(1 - F_{1})$.  This representation is free from the defects that plague
the other two representations. We have no cycles, and hence there will
be a consistent solution for any value of the exogenous variables. 
\commentout
{
In particular, there will be consistent solutions in which a fire occurs in
May: namely, $F_{1} = 1$, $C = 0$, and $F_{2} = 0$. 
}
Moreover, this model
correctly tells 
us that only an intervention on the state of the forest in May will
affect CC's camping plans. 
\commentout
{
Finally, this model allows us to express that
the later state of the forest is caused, in part, by the earlier state,
something we cannot express if there is only one variable for the state
of the forest.
} 
\exam

\commentout{
Note that it will not always be the case that choosing a representation
along one of the first two lines described above will lead to
problems. For example, if the only effect of forest fire that we are
considering is the level of tourism in August, then the first
representation may be perfectly adequate. Since a fire in May or June
will have the same impact on August tourism, we need not distinguish
between the two possibilities in our model. Similarly, it turns out that
in the model discussed by Halpern and Pearl (2000, pp. 859 - 861), the
representation along the lines of the second alternative discussed above
does no harm. In the scenario under discussion here, there is a variable
Ð CC's camping plans Ð that is causally sandwiched between the two
possible fires. It is for this reason that we must represent the state
of the forest using two separate variables, corresponding to the state
of the forest in May and June. In a different context, we might need to
discriminate between the state of the forest on even more fine-grained
time scales. 
}

Once again, our discussion of the methodology of modeling parallels
certain metaphysical discussions
in the philosophy literature.
If heavy rains 
delay the onset of a fire, is it the same fire that would have occurred
without the rains, or a different fire? 
It is hard to see how to gain traction on such an
issue by direct metaphysical speculation. By contrast, when we recast
the issue as one about what kinds of variables to include in causal
models, 
it is possible to say exactly
how the models will mislead you if you make the wrong choice. 

\section{Dealing with normality and typicality}\label{sec:final}

While the definition of causality given in Definition~\ref{actcaus}
works well in many cases, it does not always deliver answers that
agree with (most people's) intuition.  Consider the following example,
taken from Hitchcock \citeyear{Hitchcock07}, based on an example due
to Hiddleston \citeyear{Hiddleston05}.

\xam\label{xam:bogus}
Assassin is in possession of a lethal poison, but has a last-minute
change of heart and refrains from putting it in Victim's coffee.
Bodyguard puts antidote in the coffee, which would have neutralized the
poison had there been any.  Victim drinks the coffee and survives.  
Is Bodyguard's putting in the antidote a cause of Victim surviving?
Most people would say no, but
according to the preliminary HP definition, it is.  For in the contingency
where Assassin puts in the poison, Victim survives iff Bodyguard puts
in the antidote.  
\exam

Example~\ref{xam:bogus} illustrates an even deeper problem with
Definition~\ref{actcaus}.  The structural equations for
Example~\ref{xam:bogus} are \emph{isomorphic} to those in the forest-fire
example, provided that we interpret the variables appropriately.
Specifically, take the endogenous variables in Example~\ref{xam:bogus} to be
$A$ (for ``assassin does not put in poison''), $B$ (for ``bodyguard puts
in antidote''), and $\VS$ (for ``victim survives'').
Then $A$, $B$, and $\VS$ satisfy exactly the same
equations as $L$, $\ML$, and $\FF$, respectively.  In the context where
there is lightning and the arsonists drops a lit match, both 
the lightning and the match are causes of the forest fire, which seems
reasonable.  But here it does not seem reasonable that Bodyguard's
putting in the antidote is a cause.  Nevertheless, any
definition that just depends on the structural equations is bound to give
the same answers in these two examples.  (An example illustrating
the same phenomenon is given by Hall \citeyear{Hall07}.)  This suggests
that there must  
be more to causality than just the structural equations.  And, indeed, 
the final HP definition of causality allows certain contingencies to be
labeled as ``unreasonable'' or ``too farfetched''; these contingencies
are then not considered in AC2(a) or AC2(b).  As discussed by Halpern
\citeyear{Hal39}, there are problems with the HP account; we present here
the approach used in \cite{Hal39} for dealing with these problems, which 
involves assuming that  an agent has, in addition to a theory of causality
(as modeled by the structural equations), a theory of ``normality'' or
``typicality''.  (The need to consider normality was also stressed by
Hitchcock \citeyear{Hitchcock07} and Hall \citeyear{Hall07}, and further
explored by Hitchcock and Knobe \citeyear{HK09}.)  
This theory would include statements like ``typically,
people do not put poison in coffee'' and ``typically doctors do not
treat patients to whom they are not assigned''.
There are many ways of giving semantics to such
typicality statements (e.g., \cite{Adams:75,KLM,Spohn09}).  For
definiteness, we use \emph{ranking functions} \cite{Spohn09} here.

Take a \emph{world} to be a complete description of the values
of all the random variables.  
we
assume that each world has associated
with it a \emph{rank}, which is just a natural number or $\infty$.
Intuitively, the higher the rank, the less 
``normal" or ``typical"
the world.
A world with a rank of 0 is reasonably 
normal,
one with a rank of 1 is
somewhat 
normal,
one with a rank of 2 is quite 
abnormal,
and so on.
Given a ranking on worlds, the statement ``if $p$ then typically $q$''
is true if in all the worlds of least rank where $p$ is true, $q$ is
also true.  Thus, in one model where people do not typically put either
poison or antidote in coffee, the worlds where neither poison nor
antidote is put in the coffee have rank 0, worlds where either
poison or antidote is put in the coffee have rank 1, and worlds
where both poison and antidote are put in the coffee have rank 2.

Take an {\em extended causal model\/} to
be a tuple $M = (\S,\F,\kappa)$, where $(\S,\F)$ is a causal model, and
$\kappa$ is a \emph{ranking function} that associates with each world a rank.
In an acyclic extended causal model, a context $\vec{u}$ determines a
world, denoted $s_{\vec{u}}$.  
$\vec{X}=\vec{x}$ is a \emph{cause of $\phi$
in an extended model $M$ and context $\vec{u}$} if $\vec{X}=\vec{x}$ is
a cause of 
$\phi$ according to Definition~\ref{actcaus}, except that in AC2(a), 
there must be a world $s$ such that $\kappa(s) \le \kappa(s_{\vec{u}})$
and $\vec{X} = \vec{x}' \land \vec{W} = \vec{w}$ is true at $s$.
This can be viewed as a formalization of Kahneman and
Miller's \citeyear{KM86} observation that 
``an event is more likely to be undone by altering exceptional than
routine aspects of the causal chain that led to it''. 

This definition deals well with all the problematic examples in the
literature.  Consider Example~\ref{xam:bogus}.  Using
the ranking described above, Bodyguard is not a cause of Victim's
survival because the world that would need to be considered in AC2(a),
where Assassin poisons the coffee, is less normal than the actual world,
where he does not.  We consider just one other example here (see
\cite{Hal39} for further discussion).

\xam 
Consider the following story, taken from
(an early version of) \cite{Hall98}:
Suppose that Billy is hospitalized with a mild illness on Monday; he is
treated and recovers.  In the obvious causal model, the doctor's
treatment is a cause of Billy's recovery.  Moreover, if the doctor
does \emph{not} treat Billy on Monday, then the doctor's omission to
treat Billy is a cause of Billy's being sick on Tuesday.
But now suppose that there are 100
doctors in the hospital.  Although only doctor 1 is assigned to
Billy (and he forgot to give medication), in principle, any of the other
99 doctors could have given Billy his medication.  
Is the nontreatment by doctors 2--100 also a cause of Billy's being
sick on Tuesday?

\begin{sloppypar}
Suppose that in fact the hospital has 100
doctors and there are variables $A_1, \ldots, A_{100}$ and $\MT_1,
\ldots, \MT_{100}$ in the causal 
model, where $A_i = 1$ if doctor $i$ is assigned to treat Billy
and $A_i = 0$ if he is not, and
$\MT_i = 1$ if doctor $i$ actually treats Billy on Monday, and
$\MT_i = 0$ if he does not.
Doctor 1 is assigned to treat Billy; the
others are not.  However, in fact, no doctor treats Billy.  
Further assume that, typically,
no doctor is assigned to a given patient;
if doctor $i$ is not assigned to treat Billy, then typically doctor $i$
does not treat Billy;
and if doctor
$i$ \emph{is} assigned to Billy, then typically doctor $i$ treats Billy.
We can capture this in an extended causal model where the world
where no doctor is assigned to Billy and no doctor treats him has rank 0;
the 100 worlds where exactly one doctor is assigned to Billy, and that
doctor treats him, have rank
1; the 100 worlds where exactly one doctor is assigned to Billy and
no one treats him have rank 2;
and the $100\times 99$ worlds where exactly one doctor is
assigned to Billy but some 
other
doctor 
treats him have
rank 3.  (The ranking given to other
worlds is irrelevant.)  In this extended model, in the context where
doctor $i$ 
is assigned to Billy but no one treats him, $i$ is the cause of Billy's
sickness  
(the world where $i$ treats Billy 
has 
lower rank than the world where $i$ is assigned to Billy but no one
treats him), but no other doctor is a cause of Billy's sickness.
Moreover, in the context where $i$ is assigned to Billy and treats him,
then $i$ is the cause of Billy's recovery (for AC2(a), consider the world
where no doctor is assigned to Billy and none treat him).  
\end{sloppypar}
\exam

\commentout
{
Of course, once we add normality to the model, we give the modeler a
great deal of flexibility.  By choosing the normality theory
appropriately, we can model the claim that $A$ is a cause of $B$ for any
$A$ and $B$.  That makes guidelines on what counts as a reasonable model
all the more necessary.  [[MORE HERE ...]]
}
Adding a normality theory to the model gives the HP account of actual
causation greater flexibility to deal with these kinds of cases.
This raises the worry, however, that this gives the modeler too much
flexibility. After all, the modeler can now render any claim that
$A$ is an actual cause of $B$ false, simply by choosing a normality order 
that assigns the actual world $s_{\vec{u}}$ a lower rank than any world
$s$ needed to satisfy AC2. Thus, the introduction of normality exacerbates
the problem of motivating and defending a particular choice of model.
%
Fortunately, the literature on the psychology of counterfactual reasoning 
and causal judgment goes some way toward enumerating the sorts of 
factors that constitute normality. (See, for example, 
\cite{Alicke92,Cushman09,CKS08,HK09,KM86,KF08,KT82,Mendel05,Roese97}.)
\commentout{
D. Kahneman and A. Tversky, ÒThe Simulation Heuristic,Ó 
in Kahneman, D. P. Slovic, and Tversky, Judgment Under Incertainty: Heuristics and Biases 
(Cambridge: Cambridge University Press, 1982) , pp. 201 Ð 210.

D. R. Mandel, D. J. Hilton, and P. Catellani (eds.), 
The Psychology of Counterfactual Thinking (New York: Routledge, 2005). 
N. Roese, ÒCounterfactual Thinking,Ó Psychological Bulletin, CXXI (1997): 133 Ð 148.
}
These factors include the following:
\begin{itemize}
\item Statistical norms concern what happens most often, or with the
greatest frequency. 
Kahneman and Tversky \citeyear{KT82}
gave subjects a story in which
Mr. Jones usually leaves work
at 5:30, but occasionally leaves early to run errands.
Thus,  a 5:30 departure 
is (statistically) ``normal", and an earlier departure ``abnormal''. 
This difference affected which alternate possibilities subjects were
willing to consider 
when reflecting on the causes of an accident in which
Mr. Jones was involved.

\item Norms can involve moral judgments.  Cushman, Knobe, and
Sinnott-Armstrong \citeyear{CKS08}  
showed that 
people with different views about the morality of abortion
have different views about the abnormality of 
insufficient care for a fetus,
and this can lead them to make different judgments about the cause
of a miscarriage. 
\item Policies adopted by social institutions can also be norms. For
instance, Knobe and Fraser \citeyear{KF08} presented subjects with a
hypothetical situation in which a department had implemented 
a policy allowing administrative assistants to take pens from the
department office, but prohibiting faculty from doing so.
Subjects were more likely to attribute causality to a professor's
taking a pen than to an assistant's taking one, even when the situation
was otherwise similar. 
\item There can also be norms of
``proper functioning" governing the operations of biological
organs or mechanical parts: there are certain ways that hearts and 
spark plugs are ``supposed" to operate. Hitchcock and Knobe \citeyear{HK09}
show that these kinds of norms can also affect causal judgments.
\commentout
{
\item 
Normality can be defined in term of optimal performance.
In tennis, a service ace
is ``normal", while a double fault is ``abnormal" (even though the latter 
may be statistically more common, especially when novices play). 
Thus, we would typically not say in a Wimbledon match that someone won
because he hit an ace, but we would say that he lost because of a
double-fault. 
}
\end{itemize}

The law suggests a variety of principles for determining
the norms that are used in the evaluation of 
actual
causation.
In criminal law, norms are determined by direct legislation.
For example, if there are legal standards for the strength
of seat belts in an automobile, a seat belt that did not
meet this standard could be judged a cause of a traffic fatality.
By contrast, if a seat belt complied with the legal standard,
but nonetheless broke because of the extreme forces it
was subjected to during a particular accident, the fatality would
be blamed on the circumstances of the accident, rather than the
seat belt. In such a case, the manufacturers of the seat
belt would not be guilty of criminal negligence.
In contract law, compliance with the terms of a contract has
the force of a norm.
In tort law, actions are often judged against the standard of
``the reasonable person". For instance, if a bystander was
harmed when a pedestrian who was legally crossing the street
suddenly jumped out of the way of an oncoming car, the pedestrian
would not be held liable for damages to the bystander, since
he acted as the hypothetical ``reasonable person" would have done
in similar circumstances. (See, for example, \cite[pp.~142ff.]{HH85} for
discussion.) 
There are also a number of circumstances in which deliberate 
malicious acts of third parties are considered to be 
``abnormal" interventions,
and affect the assessment of causation. (See, for example,
\cite[pp.~68ff.]{HH85}.) 

As with the choice of variables, we do not expect that these 
considerations will always suffice to pick out a uniquely correct
theory of normality for a causal model. They do, however,
provide resources for a rational critique of models.

\section{Conclusion}\label{sec:conc}

As HP stress, causality is relative to a model.  That makes it
particularly important to justify whatever model is chosen, and to
enunciate principles for what makes a reasonable causal model.  We have
taken some preliminary steps in investigating this issue with regard to
the choice of variables and the choice of defaults.  However, we hope
that we have convinced the reader that far more needs to be done if
causal models are actually going to be used in applications.  

\paragraph{Acknowledgments:}  We thank Wolfgang Spohn for useful comments.
Joseph Halpern was supported in part by NSF grants 
IIS-0534064 and IIS-0812045, and by AFOSR grants
FA9550-08-1-0438 and FA9550-05-1-0055.



\nopagebreak
\end{document}